\documentclass[twoside]{article}

\usepackage{PRIMEarxiv}

\usepackage[utf8]{inputenc} 
\usepackage[T1]{fontenc}    
\usepackage{hyperref}       
\usepackage{url}            
\usepackage{booktabs}       
\usepackage{amsfonts}       
\usepackage{nicefrac}       
\usepackage{microtype}      
\usepackage{lipsum}
\usepackage{graphicx}       
\graphicspath{{media/}}     

\usepackage{graphicx, amsmath, amssymb, tabularx, booktabs, graphicx}
\usepackage{caption}
\usepackage{subcaption}
\usepackage{bm}
\usepackage{float}
\usepackage{colortbl}
\usepackage{algorithm}
\usepackage{algpseudocode}
\usepackage{amsmath}
\usepackage{xcolor}

\title{Electrostatic Force Regularization for Neural Structured Pruning
\thanks{This paper is under review at CVPR 2025.}
}

\author{
  Abdesselam Ferdi \\
  Signal Processing Laboratory \\
  Constantine 1 - Frères Mentouri University \\
  Constantine, Algeria\\
  \texttt{abdesselam.ferdi@gmail.com} \\
   \And
   Abdelmalik Taleb-Ahmed \\
   Université Polytechnique Hauts-de-France \\
   Université de Lille, CNRS \\
   Valenciennes, France \\
   \texttt{abdelmalik.taleb-ahmed@uphf.fr}
   \And
Amir Nakib \\
Université Paris Est Créteil, Laboratoire LISSI, \\ 122 Rue Paul Armangot 94400 \\ Vitry Sur Seine, France \\
\texttt{nakib@u-pec.fr}
     \And
   Youcef Ferdi \\
   Bioengineering Laboratory \\
   National Higher School of Biotechnology \\
   Constantine, Algeria \\
   \texttt{y.ferdi@ensbiotech.edu.dz}
}

\begin{document}
\maketitle

\begin{abstract}
The demand for deploying deep convolutional neural networks (DCNNs) on resource-constrained devices for real-time applications remains substantial. However, existing state-of-the-art structured pruning methods often involve intricate implementations, require modifications to the original network architectures, and necessitate an extensive fine-tuning phase. To overcome these challenges, we propose a novel method that, for the first time, incorporates the concepts of charge and electrostatic force from physics into the training process of DCNNs. The magnitude of this force is directly proportional to the product of the charges of the convolution filter and the source filter, and inversely proportional to the square of the distance between them. We applied this electrostatic-like force to the convolution filters, either attracting filters with opposite charges toward non-zero weights or repelling filters with like charges toward zero weights. Consequently, filters subject to repulsive forces have their weights reduced to zero, enabling their removal, while the attractive forces preserve filters with significant weights that retain information. Unlike conventional methods, our approach is straightforward to implement, does not require any architectural modifications, and simultaneously optimizes weights and ranks filter importance, all without the need for extensive fine-tuning. We validated the efficacy of our method on modern DCNN architectures using the MNIST, CIFAR, and ImageNet datasets, achieving competitive performance compared to existing structured pruning approaches.
\end{abstract}

\keywords{DCNNs \and Electrostatic force \and Structured pruning \and Regularization}

\section{Introduction}
With the rise of deep learning models, which are often characterized by their memory requirements (number of parameters) and computational power (floating point operations per second, or FLOPs), there is a growing demand for model acceleration techniques. These techniques are designed to convert complex, resource-intensive models into more efficient, lightweight versions. This transformation enables the deployment of deep learning models on devices with limited resources, making them more accessible and practical in a variety of settings. Model compression is a subset of the model acceleration field. It encompasses a range of techniques, including distillation \cite{zheng2024restructuring}, low-rank approximation \cite{song2024low}, quantization \cite{pei2023quantization}, and pruning \cite{liu2021discrimination}. Pruning itself can be further divided into unstructured, semi-structured, and structured methods. Recent research has primarily focused on structured pruning (SP), which forms the foundation of our work. In this context, entire filters or channels are completely removed from the convolutional layer, leading to a reduction in the model's parameters and FLOPs \cite{10398587}.

The key question we address in this work is: How can a model be optimally configured for the pruning stage to achieve minimal accuracy drop?

Our work proposes a novel, physics-inspired approach for SP of deep convolutional neural networks (DCNNs). To the best of our knowledge, this paper introduces, for the first time, the concepts of \textit{charges} and \textit{electrostatic force} from physics into DCNNs. This electrostatic force is integrated into the training process of DCNNs, exerting an attractive or repulsive force on the filters of the convolutional layer. Depending on the charge of the filter, filters dissimilar to the source filter experience an attractive force, moving their weights toward non-zero weights. Conversely, similar filters experience a repulsive force, moving their weights toward zero. The term 'source filter' refers to the filter with the largest magnitude, computed using the $L_1$-norm. In essence, we induced sparsity in the convolutional layer by applying the repulsive force. The magnitude of the electrostatic force is directly proportional to the product of the magnitudes of the charge of the source filter and the filter, and inversely proportional to the square of the distance between them. Upon completion of model training with the electrostatic force and based on a predetermined pruning ratio, we can prune filters that experience a repulsion force, which represent less important weights, from the model, while preserving the information from the attractive filters, which represent more important weights. Our proposed method reduces the model's complexity with minimal accuracy drop. It does not require any modifications to the model architecture and allows for pruning at varying pruning ratios without the need for retraining. This feature sets our method apart from existing SP methods, which typically train the model for a predefined pruning ratio and require retraining when this ratio is changed.

Our contributions are as follows: 
\begin{itemize}
    \item We introduce, for the first time, the concepts of charge and electrostatic force from physics into the training stage of DCNNs.
    \item  We propose a novel, physics-inspired SP method for DCNNs, analogous to the concept of electrostatic force from physics.
    \item Our method effectively reduces the model's complexity in terms of both memory requirements and energy consumption.
    \item Our method is easy to implement and does not necessitate any modifications to the model architecture. It can be applied to both non-trained and pretrained models.
    \item We can prune a model trained with our method at varying pruning ratios without the need for retraining.
\end{itemize} 

In the rest of this paper. In Section 2, we review the state-of-the-art (SOTA) methods for DCNNs pruning. In Section 3, we delve into our methodology, discussing both the motivation behind our work and the specifics of our proposed method. In Section 4, we detail the experimental setup used to validate our method. In Section 5, we present the results obtained from our experiments. Finally, in Section 6, we conclude the work and discuss potential avenues for future research. 

\section{Related Work}
The advancement of complex and deep models continues to show promising results, making model acceleration a vibrant research field. Over the last ten years, model pruning has gained significant attention due to the high demand for pruned models in the practical application of deep networks. Model pruning is classified into three categories based on its incorporation during network training: pruning at initialization, during training, and after training. These categories include three methods: semi-structured \cite{Grimaldi_2023_ICCV}, unstructured \cite{zhang2022advancing}, and SP \cite{cai2022prior}. Our work is particularly focused on SP after network training.

Pruning at initialization approaches, inspired by the Lottery Ticket Hypothesis \cite{frankle2018lottery}, challenge the necessity of dense training for achieving performance convergence \cite{liu2018rethinking}. A pioneering approach in this category is SNIP \cite{lee2018snip}, which identifies a trainable sub-network at initialization. Following this, methods like FORCE \cite{de2020progressive} and GraSP \cite{wang2020picking} have been introduced to further enhance performance. These approaches enhance training efficiency by focusing solely on the sparse network. However, the reliability of pruning at initialization remains suboptimal, as it often leads to unavoidable performance gaps \cite{frankle2020linear}.

Pruning during training approaches remove unimportant parameters during model training. Existing works can be categorized into two main methods: regularization-based methods, which promote sparsity during training \cite{wang2021neural,zhang2022advancing,Yang_2023_CVPR}, and sub-ticket selection methods, which use saliency measures to eliminate redundant components \cite{aketi2020gradual,he2020learning,frankle2020linear,oyedotun2020structured}. These methods face challenges in automatically determining the optimal starting points for pruning, often relying heavily on hand-crafted rules or post-training heuristics to guide the decision-making process.

Post-training pruning methods involve pruning a densely pretrained model followed by fine-tuning to recover any information loss due to pruning \cite{evci2020rigging,cai2022prior,xu2023efficient,gupta2024torque}. These approaches aim to identify redundant connections whose removal has the least impact on overall performance. Although they deliver competitive results and improve test-time efficiency, they fail to enhance training efficiency. In fact, many of these techniques nearly double the training time due to the need for extensive fine-tuning stage.

\section{Approach}
This section begins by discussing the motivation for applying the concept of electrostatic force from physics to the SP of DCNNs, followed by a detailed explanation of the proposed method.

\subsection{Motivation}
In nature, matter possesses two fundamental properties: mass and charge. The latter property, charge, causes matter to experience a force when it is placed within an electrostatic field. The charge of an object is determined by the number of protons and electrons it contains. An object will have a positive charge if it has more protons than electrons. Conversely, if an object has more electrons than protons, it will carry a negative charge. If the number of protons and electrons is equal, the object is considered electrically neutral. The interaction between charges can be either repulsion or attraction. Like charges repel each other, while unlike charges attract. This principle was formulated into Coulomb’s law by physicist \textit{Charles-Augustin de Coulomb} in 1785, which describes the electrostatic force between stationary charged objects.

We propose the integration of the concept of electrostatic force from physics into the training stage of DCNNs. Our hypothesis is that at each convolutional layer, filters with the same sign (positive or negative) as the source filter will tend toward zero weights due to repulsion, while filters with the opposite sign will gravitate toward non-zero weights due to attraction. Subsequently, filters with zero weights, which contribute minimal information, could be pruned with minimal loss of information, while preserving the more important information from the filters with non-zero weights. We believe that this approach, drawing parallels with electrostatic interactions, will significantly enhance the efficiency of DCNNs.

\subsection{Proposed Method}
Coulomb's law states that the magnitude of the electrostatic force between two charges, denoted as $q_1$ and $q_n$, is 
\begin{equation}
   F_e = k_e \frac{|q_1| |q_n|}{r^2}
   \label{eq1}
\end{equation}

where $k_e$ denotes the Coulomb constant (8.99$\times 10^9 Nm^2C^{-2}$) and $r$ represents the distance between the charges. 

Similar to the Coulomb's law, we define the magnitude of the electrostatic force between two charges of filters in a DCNN by Eq. \ref{eq1}, where $q_1$ and $q_n$ represent the charges (or their magnitudes, $|q_1|$ and $|q_n|$) corresponding to the source filter and filter $n$, respectively. We denote $k_e$ as the Coulomb constant and $r$ as the distance between the source filter and filter $n$.

To understand how the concept of the electrostatic force has been applied in the context of DCNNs pruning, let us consider a convolutional layer $l$ with $N$ filters, each of shape $[c, k, k]$, where $c$ and $k$ denote the number of input channels and filter shape, respectively. At each layer $l$, a source filter will create an electrostatic field that applies either: (1) Attractive force on dissimilar filters (i.e., signs of the charges of the source filter and the filter are different). (2) Repulsive force on like filters (i.e., signs of the charges of the source filter and the filter are the same). (3) No force on itself and neutral filters (i.e., sign of the charge of the filter is zero). 

We define the charge $q_{n,l}$ of filter $n$ of the convolutional layer $l$ by its sign and magnitude as follows:
\begin{align}
  q_{n,l} &= (q_{n,l}) \times |q_{n,l}| \notag \\
  &= (n_l) \times |n_l|
  \label{eq3}
\end{align}

where $n \in [0, N-1]$. The function $(.)$ is the sign function and $|.|$ denotes the magnitude. 

We define the magnitude of filter $n$ as the L$_1$-norm of its weights, that is
\begin{equation}
    |n_l| = \|W_{n,l}\|_1
\end{equation}

The weights of the filters in a DCNN are floating-point numbers that can be both positive and negative. In light of this property, we define the sign of the filter as
\begin{equation}
  (n_l) = (\sum_{i=1}^{ckk} (w_i))
  \label{eq4}
\end{equation}

where $w_i \in \mathbb{R}^{1 \times ckk}$ represents the reshaped weights of the filter.

We define the distance between the source filter and filter $n$ in terms of charges as follows
\begin{equation}
  r_{n,l} = |q_{1,l} - q_{n,l}|
  \label{eq5}
\end{equation}

We generalize Eq. \ref{eq1} for the $n^{th}$ filter of layer $l$ as
\begin{equation}
  F_{e_{n,l}} = k_e \frac{|q_{1,l}| |q_{n,l}|}{r_{n,l}^2} 
  \label{eq6}
\end{equation}

Upon examining Eq. \ref{eq5}, we can discern two distinct cases: (1) When the charge of filter $n$ $(q_{n,l})$ has the same sign as the charge of the source filter $(q_{1,l})$, the distance $r_{n,l}$ decreases. (2) Conversely, if the charges bear different signs, the distance increases. Consequently, a smaller distance corresponds to a stronger electrostatic force $F_{e_{n,l}}$, while a greater distance results in a weaker electrostatic force. 

It’s important to note, however, that the electrostatic force ceases to exist in the following two cases: (1) When the filter is neutral and (2) when considering the source filter interacts with itself.

We regularize the cost function with the electrostatic force as follows
\begin{equation}
    \Tilde{J} = \sum_{n,l} J(w_{n,l}; X, y) + \alpha_e F_{e_{n,l}}
    \label{eq7}
\end{equation}

where $\alpha_e$ represents the electrostatic force rate that weights the relative contribution of the penalty term, $F_{e_{n,l}}$, relative to the standard objective function $J$.

\begin{figure}[t]
    \centering
    \includegraphics[height=0.5\textwidth, width=0.8\linewidth]{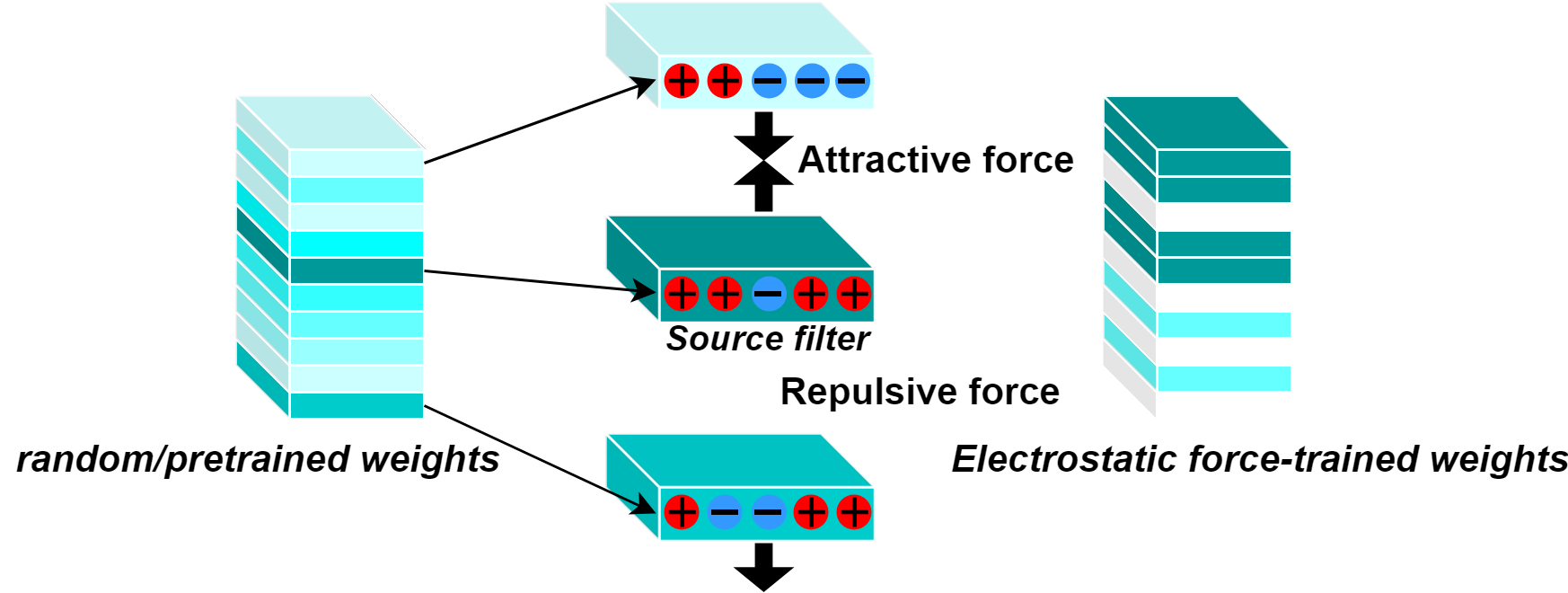}
    \caption{An illustration of electrostatic force-based training. A convolutional layer with ten filters is used as an example. The color shading represents the magnitude of each filter, with the source filter exhibiting the largest magnitude. The signs (plus and minus) indicate the polarity of the filter weights, corresponding to the filter charge (positive or negative). Initially, the convolutional layer contains filters with either random or pretrained weights. Filters with charges similar to the source filter (i.e., positive charges) experience repulsive forces, pushing their weights toward zero. Conversely, filters with opposite charges experience attractive forces, pulling their weights toward non-zero values.}
  \label{figure1}
\end{figure}

When the optimizer minimizes the regularized objective function $\Tilde{J}$, it will decrease both the standard objective function $J$ on the training data ($X, y$) and the magnitude of the electrostatic force $F_{e_{n,l}}$. The act of minimizing the $F_{e_{n,l}}$ influences the weights of filters. Specifically, filters that are subjected to a repulsive force (i.e., larger electrostatic force) are driven toward zero weights, while those experiencing an attractive force (i.e., smaller electrostatic force) are guided toward non-zero weights. The influence of the electrostatic force on filter weights is illustrated in Figure \ref{figure1}.

The parameter gradient of the regularized objective function with respect to the weights is
\begin{equation}
    \nabla \Tilde{J} = \sum_{n,l} \nabla_{w_{n,l}} J(w_{n,l}; X, y) + \alpha_e \nabla_{w_{n,l}} F_{e_{n,l}}
    \label{eq8}
\end{equation}

By substituting Eq. \ref{eq6} into Eq. \ref{eq8}  (we disregard $n$ and $l$), we can write
\begin{equation}
    \nabla \Tilde{J} = \nabla_w J(w; X, y) + \alpha_e \nabla_w k_e \frac{|q_1| |q|}{r^2}
    \label{eq9}
\end{equation}

Thus
\begin{equation}
    \nabla \Tilde{J} = \nabla_w J(w; X, y) + \alpha_e k_e \frac{|q_1|}{r^2} (w)
    \label{eq10}
\end{equation}

The weights are updated by a single gradient step using the following weight-update rule
\begin{equation}
    w \leftarrow w - \epsilon(\nabla_w J(w; X, y) + \alpha_e k_e \frac{|q_1|}{r^2} (w))
    \label{eq11}
\end{equation}

where $\epsilon$ is the learning rate.

By inspecting Eq. \ref{eq11}, we see that the influence of the electrostatic force on the gradient is a factor of the same sign as $w$. This factor is primarily dependent on the square of the distance, denoted as $r$, between the source filter and filter $n$. Consequently, for smaller distances, the gradient is amplified, corresponding to a repulsive force. In this case, the optimization algorithm prompts the filter weights to approach zero. Conversely, for larger distances, the gradient diminishes, corresponding to an attractive force. Here, the filter weights are updated to gravitate toward non-zero values.

To facilitate understanding, we summarized our proposed method in Algorithm \ref{algorithm1}. It is important to highlight that we applied the electrostatic force only to the convolutional layers to prune in the pruning phase. The rationale for this rule is: When the optimizer minimizes the regularized objective function, denoted as $\Tilde{J}$, it concurrently minimizes both the standard objective function $J$ and the electrostatic force $F_e$. The minimization of the standard objective function nudges the filter weights toward local minima. Conversely, the minimization of the electrostatic force guides the weights of the filters to prune toward a specific weight distribution, facilitating their removal with low information loss. 

\begin{algorithm}[t]
\caption{Electrostatic force-based training algorithm}
\label{algorithm1}
\begin{algorithmic}[1]
\State \textbf{Input:} Randomly initialized or pretrained model $M$, convolutional layers to prune $L$, Coulomb constant $k_e$, electrostatic force rate $\alpha_e$, epochs $E$
\State \textbf{Output:} Electrostatic force-trained model $M_{F_e}$
\For{layer $l$ in $L$}
    \State Extract and flatten the weights of filters of layer $l$
    \For{filter $n$ in layer $l$}
        \State Compute the magnitude $|n_l|$ (based on L$_1$-norm of weights) and the sign $(n_l)$ (based on the signs of weights) of the filter $n$
        \State Compute the charge $q_{n,l}$ of the filter $n$
        \State Rank the magnitudes of filters and determine the source filter $(q_{1,l})$ (with biggest magnitude)
        \State Compute the distances $r_{n,l}$ between convolution filter $n$ and the source filter (in terms of charges)
        \If{$n = source\_filter\_index$ \textbf{or} $(n_l) = 0$}
            \State $F_{e_{n,l}} = 0$
        \Else
            \State $F_{e_{n,l}} = k_e \frac{|q_{1,l}| |q_{n,l}|}{r_{n,l}^2}$
        \EndIf
    \EndFor
\EndFor
\State Compute the sum of the magnitudes of the electrostatic forces $F_e$ for the model $M$
\State Regularize the standard objective function $J$ based on the $\alpha_e$
\State Train the model $M$ for $E$ epochs using stochastic gradient descent (SGD) with the regularized objective function $\Tilde{J}$
\end{algorithmic}
\end{algorithm}

\section{Experimental Setup}
This section provides details on our experimental design, including the datasets, networks, implementation procedure, and experimental setting.

\subsection{Datasets and Networks}
We first conducted analyses on the MNIST \cite{geng2024orthcaps} and CIFAR \cite{iurada2024finding} datasets with ResNet \cite{gao2024bilevelpruning} and VGGNet \cite{xie2024unipts}. Subsequently, we continue with further analyses on the large-scale ImageNet dataset \cite{iurada2024finding} with ResNets. In our experiments, we deliberately excluded VGGNet from the ImageNet benchmark. This is due to their single-branch architecture, which no longer reflects the structure of modern deep networks with residual connections.

\subsection{Implementation}
In this section, we present a detailed description of the procedure we employed to implement our method. We begin by clarifying the network training with our method, which enables us to compress the filters at each convolutional layer within the network. Following this, we describe the pruning strategy that we used to remove the compressed filters from the model.

\subsubsection{Electrostatic Force-based Training}
We defined the magnitude of the electrostatic force in DCNNs as the product of the constant $k_e$ and the magnitude of charges $q_1$ and $q_n$, all divided by the square of the distance between them. The charges of the source filters and the distances were computed once at each convolutional layer within the model. The model's objective function is influenced by the electrostatic force, which is weighted by the hyperparameter $\alpha_e$ (as per Eq. \ref{eq7}). In our analyses, we compared electrostatic force-based training of models initialized with random weights versus pretrained weights.

The model's parameters were optimized using the stochastic gradient descent optimizer. We employed the PyTorch 1.9.0 framework for implementation with the Anaconda software on an NVIDIA T600 GPU. The operating system underpinning our experimental framework is GNU/Linux x86\textunderscore64. In Table \ref{table1}, we provide a comprehensive summary of the training settings, outlining the key parameters and configurations used throughout the experiments.
\begin{table}[b]
  \caption{Summary of the electrostatic force-based training setting. The term "pr" describes the settings used in model training without fine-tuning. The term "ft" refers to the settings used in the fine-tuning of the pruned model. The term "pr + ft" refers to the settings used in the training and fine-tuning of the model. For the SGD optimizer, in the parentheses are the momentum and weight decay. For the LR policy, "P1" refers to a multi-step: (0: 1e-1, 100: 1e-2, 150: 1e-3) and "P2" refers to a multi-step: (0: 1e-2, 60: 1e-3, 90: 1e-4).}
  \centering
    \begin{tabularx}{\textwidth}{>{\centering\arraybackslash}X||>{\centering\arraybackslash}X|>{\centering\arraybackslash}X|>{\centering\arraybackslash}X}
    \hline
    Dataset & MNIST & CIFAR & ImageNet \\ 
    \hline
    \hline
    Optimizer (pr) & SGD($0.9$, $0$) & SGD($0.9$, $0$) & - \\
    Optimizer (pr + ft) & - & SGD($0.9$, $5$e-$4$) & SGD($0.9$, $5$e-$4$) \\
    LR policy (pr) & P1 & P1 & - \\
    LR policy (ft) & - & P2 & P2 \\
    Total epoch (pr) & $200$ & $200$ & $100$ \\
    Total epoch (ft) & - & $120$ & $80$ \\ 
    Batch size & $256$ & $128$ & $512$ \\ 
    \hline
    \end{tabularx}
    \label{table1}
\end{table}

\subsubsection{Pruning}
After completing the electrostatic force-based model training, we proceed to the pruning phase. This phase involves eliminating filters with zero weights (or less important weights) from the model, thereby ensuring minimal loss of information. We adopted a local pruning strategy to identify filters for removal at each convolutional layer within the model. This strategy involves ranking the filters based on their L$_1$-norm and then removing those with the smallest norm based on a consistent pruning rate across all convolutional layers. In Table \ref{table2}, we list the electrostatic force rates, denoted as $\alpha_e$, employed in each of our experiments. Furthermore, we present the specific pruning ratios used, along with the corresponding speedups achieved for each configuration. Our approach to pruning adheres to established practices \cite{wang2021neural}. (1) For a ResNet model comprising N stages, the pruning ratio is represented as a list of N floats. To illustrate, a ResNet-56 model comprising five stages is pruned with a ratio of [0, 0.52, 0.52, 0.52, 0]. This implies that the initial convolutional layer, which constitutes the first stage, has a pruning ratio of 0; the subsequent two-convolutional layer bottleneck blocks have a pruning ratio of 0.52; and the final stage, which is the fully connected layer, has a pruning ratio of 0. It should be noted that the last convolution in a bottleneck block is not pruned. This implies that for a two-convolutional layer bottleneck block (as in ResNet-56), only the initial layer is subjected to pruning. Similarly, for a three-convolutional layer bottleneck block (as in ResNet-34 and 50), only the initial two layers are pruned. (2) For a VGG19 model, a pruning ratio of [0:0, 1-15:0.65] indicates that the first convolutional layer has a pruning ratio of 0, while convolutional layers 1 to 15 have a pruning ratio of 0.65.

\begin{table}[t]
    \caption{Summary of the electrostatic force rate and pruning ratio.}
    \centering
    \begin{tabularx}{\textwidth}{>{\centering\arraybackslash}X||>{\centering\arraybackslash}X|>{\centering\arraybackslash}X|>{\centering\arraybackslash}X}
    \hline
    Model/Dataset & $\alpha_e$ & Pruning ratio & Speedup \\ 
    \hline
    \hline
    ResNet-56/CIFAR-10 & $10^{-16}$ & $[0, 0.52, 0.52, 0.52, 0]$ & $2.17\times$\\ 
    ResNet-56/CIFAR-10 & $10^{-16}$ & $[0, 0.6, 0.6, 0.6, 0]$ & $2.62\times$\\ 
    ResNet-56/CIFAR-10 & $10^{-16}$ & $[0, 0.62, 0.63, 0.62, 0]$  & $2.73\times$ \\ 
    VGG-19/CIFAR-100 & $10^{-11}$ &  $[0:0, 1-15:0.65]$ & $6.85\times$ \\ 
    VGG-19/CIFAR-100 & $10^{-11}$ & $[0:0, 1-15:0.70]$ & $8.89\times$ \\ 
    ResNet-34/ImageNet & $10^{-17}$ & $[0, 0.5, 0.5, 0.5, 0, 0]$ & $1.34\times$\\ \hline
    \end{tabularx}
    \label{table2}
\end{table}

\subsubsection{Fine-Tuning}
Our electrostatic force-based DCNNs training method shows promising results in pruning. For a fair comparison with SOTA SP methods, we fine-tuned our pruned models. We used the same scheme for this phase, which includes hyperparameters like number of epochs, learning rate, batch size, etc., as employed by other pruning methods \cite{wang2021neural}.

\section{Results}
In this section, we present the pruning results obtained with the proposed method, both before and after fine-tuning. Additionally, we report the training cost associated with the proposed method.

\subsection{Pruning Results}
\begin{figure*}[b]
  \centering
  \includegraphics[width=\linewidth]{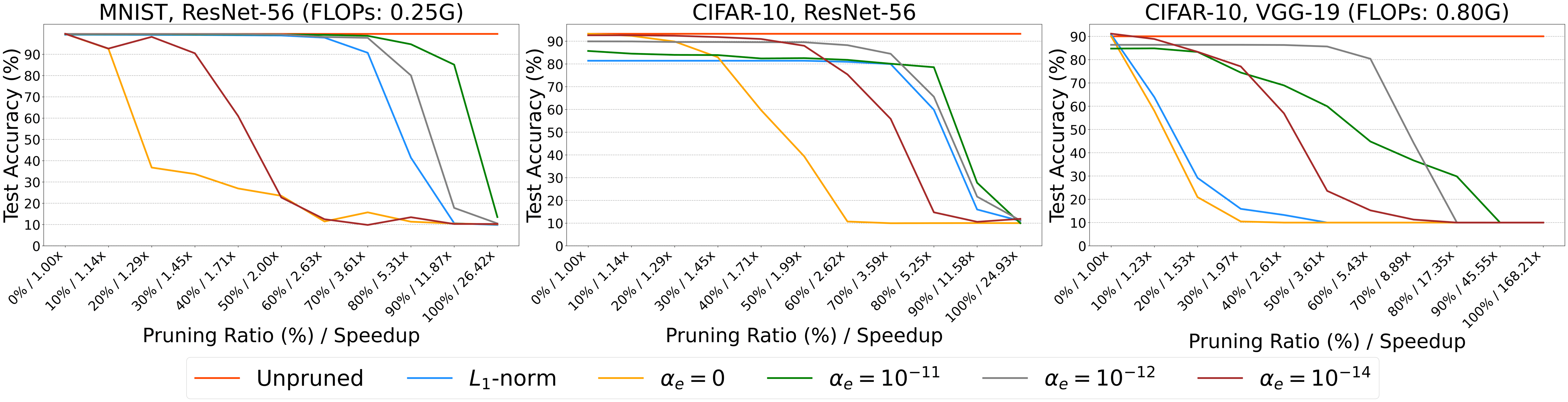}
  \caption{Top-1 accuracy of pruned ResNet-56 and VGG-19 models, initialized with pretrained weights and trained with L$_1$-norm and electrostatic force at four distinct electrostatic force rates, without any fine-tuning.}
  \label{figure2}
\end{figure*}

We trained the networks using the electrostatic force with pretrained weights on the MNIST and CIFAR datasets. Our baseline models were trained on the same datasets with accuracies comparable to those reported in the original papers. Furthermore, we trained the same networks with L$_1$-norm at a regularization rate of $10^{-2}$. We present the pruning results obtained before fine-tuning in the form of curves in Figure \ref{figure2}. For each network, we plotted the pruned top-1 accuracy versus the pruning ratio (and the corresponding speedup rate), which varies in the range of $[0\%, 100\%]$.

\subsection{Training Overhead}
We calculated the training cost without fine-tuning for the ResNet-56 and VGG-19 models on the MNIST and CIFAR-10 datasets, and the results are presented in Table \ref{table3}.

\begin{table}[b]
  \caption{Training cost comparison of ResNet-56 and VGG-19 models trained with the baseline method, L$_1$-norm, and electrostatic force method on an NVIDIA T600 GPU.}
  \centering
    \begin{tabularx}{\textwidth}{>{\centering\arraybackslash}X||>{\centering\arraybackslash}X|>{\centering\arraybackslash}X}
    \hline
    Model/Dataset & Method & Training time (h)\\ 
    \hline
    \hline
    & Baseline & 1.23 \\
    ResNet-56/MNIST & L$_1$-norm & 4.34 \\
    & \cellcolor{gray!30} Electrostatic force (\textbf{ours}) & \cellcolor{gray!30} 5.41 \\ 
    \hline
    & Baseline & 1.05 \\
    ResNet-56/CIFAR & L$_1$-norm & 3.05 \\
    & \cellcolor{gray!30} Electrostatic force (\textbf{ours}) & \cellcolor{gray!30} 4.91 \\ 
    \hline
    & Baseline & 0.73 \\
    VGG-19/CIFAR & L$_1$-norm & 7.42 \\
    & \cellcolor{gray!30} Electrostatic force (\textbf{ours}) & \cellcolor{gray!30} 12.06 \\ 
    \hline    
    \end{tabularx}
    \label{table3}
\end{table}

\subsection{Fine-Tuning Results}
We compared our pruned electrostatic force-trained models against existing pruned models, with the results presented in Tables \ref{table4}, \ref{table5}, and \ref{table6}. For SOTA SP methods, we presented the results as reported in the original papers. 

We first evaluate our method on the CIFAR dataset with ResNet and VGGNet, as presented in Tables \ref{table4} and \ref{table5}. Methods with similar speedup are grouped together for easy comparison. Then we compare our method with existing methods on the standard large-scale ImageNet benchmark with ResNet. The results are presented in Table \ref{table6}. 
\begin{table}[t]
\caption{Comparison of different methods on CIFAR-10 with ResNet-56. \textit{r} and \textit{p} correspond to an electrostatic force-trained model with randomly initialized and pretrained weights, respectively. A negative value in \textit{Acc. drop} indicates an improved model accuracy. For details on the speedup configuration, please refer to Table \ref{table2}.}
\label{table4}
\begin{tabularx}{\linewidth}{>{\centering\arraybackslash}p{4cm}||>{\centering\arraybackslash}X|>{\centering\arraybackslash}X|>{\centering\arraybackslash}X|>{\centering\arraybackslash}X}
\hline
Method & Base & Pruned & Acc. & Speed \\ 
& acc. (\%) & acc. (\%) & drop (\%) & up \\
\hline
\hline
GReg-1 \cite{wang2021neural} & $93.51$ & $93.25$ & $0.26$ & $1.99\times$ \\
GReg-2 \cite{wang2021neural} & $93.51$ & $93.28$ & $0.23$ & $1.99\times$ \\ 
CP \cite{he2017channel} & $92.80$ & $91.80$ & $1.00$ & $2.00\times$ \\
AMC \cite{he2018amc} & $92.80$ & $91.90$ & $0.90$ & $2.00\times$ \\
FPGM \cite{he2019filter} & $93.59$ & $93.26$ & $0.33$ & $2.11\times$ \\
SFP \cite{he2018soft} & $93.59$ & $93.36$ & $0.23$ & $2.11\times$ \\
WHC \cite{chen2023whc} & $93.59$ & $93.74$ & $0.15$ & $2.11\times$ \\
LFPC \cite{he2020learning} & $93.59$ & $93.24$ & $0.35$ & $2.12\times$ \\
Torque \cite{gupta2024torque} & $93.48$ & $93.76$ & $-0.28$ & $2.15\times$ \\
\rowcolor{gray!30}
Electrostatic force (r) (\textbf{ours}) & $93.77$ & $93.45$ & $0.32$ & $2.17\times$ \\
\rowcolor{gray!30}
Electrostatic force (p) (\textbf{ours}) & $94.05$ & $93.88$ & $0.17$ & $2.17\times$ \\
\hline
ABC Pruner \cite{lin2020channel} & $93.26$ & $93.23$ & $0.03$ & $2.18\times$ \\
RL-MCTS \cite{wang2022channel} & $93.20$ & $93.56$ & $-0.36$ & $2.22\times$ \\
C-SGD \cite{ding2019centripetal} & $93.39$ & $93.44$ & $-0.05$ & $2.55\times$ \\ 
GReg-1 \cite{wang2021neural} & $93.51$ & $93.18$ & $0.18$ & $2.55\times$ \\
GReg-2 \cite{wang2021neural} & $93.51$ & $93.36$ & $0.00$ & $2.55\times$ \\
AFP \cite{ding2018auto} & $93.93$ & $92.94$ & $0.99$ & $2.56\times$ \\
Torque \cite{gupta2024torque} & $93.48$ & $93.40$ & $0.08$ & $2.60\times$ \\
\rowcolor{gray!30}
Electrostatic force (r) (\textbf{ours}) & $93.77$ & $93.04$ & $0.73$ & $2.62\times$ \\
\rowcolor{gray!30}
Electrostatic force (p) (\textbf{ours}) & $94.05$ & $93.64$ & $0.41$ & $2.62\times$ \\
\hline
WHC \cite{chen2023whc} & $93.59$ & $93.29$ & $0.30$ & $2.71\times$ \\
Torque \cite{gupta2024torque} & $93.48$ & $93.26$ & $0.22$ & $2.72\times$ \\
\rowcolor{gray!30}
Electrostatic force (r) (\textbf{ours}) & $93.77$ & $93.10$ & $0.67$ & $2.73\times$ \\
\rowcolor{gray!30}
Electrostatic force (p) (\textbf{ours}) & $94.05$ & $93.57$ & $0.48$ & $2.73\times$ \\
\hline
\end{tabularx}
\end{table}

\begin{table}[t]
\caption{Comparison of different methods on CIFAR-100 with VGG-19. \#Parameters: 20.07M, FLOPs: 0.80G.}
\label{table5}
\begin{tabularx}{\linewidth}{>{\centering\arraybackslash}p{4cm}||>{\centering\arraybackslash}X|>{\centering\arraybackslash}X|>{\centering\arraybackslash}X|>{\centering\arraybackslash}X}
\hline
Method & Base & Pruned & Acc. & Speed \\ 
& acc. (\%) & acc. (\%) & drop (\%) & up \\
\hline
\hline
Kron-OBD \cite{wang2019eigendamage} & $73.34$ & $60.70$ & $12.64$ & $5.73\times$ \\
Kron-OBS \cite{wang2019eigendamage} & $73.34$ & $60.66$ & $12.68$ & $6.09\times$ \\
\rowcolor{gray!30}
Electrostatic force (r) (\textbf{ours}) & $74.38$ & $68.32$ & $6.06$ & $6.85\times$ \\
\rowcolor{gray!30}
Electrostatic force (p) (\textbf{ours}) & $74.59$ & $69.00$ & $5.59$ & $6.85\times$ \\
\hline
EigenDamage \cite{wang2019eigendamage} & $73.34$ & $65.18$ & $8.16$ & $8.80\times$ \\
GReg-1 \cite{wang2021neural} & $74.02$ & $67.55$ & $6.47$ & $8.84\times$ \\
GReg-2 \cite{wang2021neural} & $74.02$ & $67.75$ & $6.27$ & $8.84\times$ \\
Torque \cite{gupta2024torque} & $73.03$ & $65.87$ & $7.16$ & $8.88\times$ \\
\rowcolor{gray!30}
Electrostatic force (r) (\textbf{ours}) & $74.38$ & $66.72$ & $7.66$ & $8.89\times$ \\
\rowcolor{gray!30}
Electrostatic force (p) (\textbf{ours}) & $74.59$ & $67.53$ & $7.06$ & $8.89\times$ \\
\hline
\end{tabularx}
\end{table}

\begin{table}[t]
\caption{Acceleration comparison on ImageNet with ResNet-34. FLOPs: 3.66G.}
\label{table6}
\begin{tabularx}{\linewidth}{>{\centering\arraybackslash}p{4cm}||>{\centering\arraybackslash}X|>{\centering\arraybackslash}X|>{\centering\arraybackslash}X|>{\centering\arraybackslash}X}
\hline
Method & Base & Pruned & Acc. & Speed \\ 
& acc. (\%) & acc. (\%) & drop (\%) & up \\
\hline
\hline
Taylor-FO \cite{molchanov2019importance} & $73.31$ & $72.83$ & $0.48$ & $1.29\times$ \\
$L_1$ (pruned-B) \cite{li2016pruning} & $73.23$ & $72.17$ & $1.06$ & $1.32\times$ \\
GReg-1 \cite{wang2021neural} & $73.31$ & $73.54$ & $-0.23$ & $1.32\times$ \\
GReg-2 \cite{wang2021neural} & $73.31$ & $73.61$ & $-0.30$ & $1.32\times$ \\    
\rowcolor{gray!30}
Electrostatic force (p) (\textbf{ours}) & $73.91$ & $73.72$ & $0.19$ & $1.34\times$ \\
\hline
\end{tabularx}
\end{table}

\section{Discussion}
In this section, we begin by analyzing the results obtained prior to fine-tuning, followed by a discussion of the results after the fine-tuning stage. Additionally, we examine the influence of the hyperparameters on the performance of the proposed method.

\subsection{Comparison With L1-norm}
Due to limitations in computing resources, we reported only the classification results of the ResNet-56 and VGG-19 models on the MNIST and CIFAR datasets. In our experiments with: (1) The ResNet-56/MNIST model, accelerated with speedup rates of 5.31$\times$ and 11.87$\times$, exhibits accuracy drops of 4.88\% and 14.46\%, respectively, when training with the electrostatic force at $\alpha_e = 10^{-11}$. However, in the case of L$_1$-norm-based training, these accuracy drops are increased by 53.3\% and 74,49\%. On the CIFAR-10 dataset, with speedup rates of 3.59$\times$ and 5.25$\times$, we observe a more substantial accuracy drops of 8.48\% and 27.73\%, respectively, when using the electrostatic force at $\alpha_e = 10^{-12}$. In the case of L$_1$-norm-based training, the aforementioned accuracy drops are increased by 4.5\% and 5.72\%, respectively. (2) The VGG-19/CIFAR-10 model, exhibits a 4.4\% and 9.68\% reduction in accuracy, at 3.61$\times$ and 5.53$\times$ speedup rates, respectively, when training with the electrostatic force at $\alpha_e = 10^{-12}$. However, with L$_1$-norm-based training, we lost almost entirely accuracy (80.06\% drop) at the same speedup rates.

In the context of training time, Table \ref{table3} illustrates that the baseline method yields the most efficient training durations across both models and datasets. In contrast, the L$_1$-norm and electrostatic force methods result in substantial increases in training time. Notably, although the electrostatic force method demonstrates improved pruning results, it incurs a greater training overhead, particularly for the VGG-19 model on the CIFAR-10 dataset.

\begin{figure*}[t]
  \centering
\begin{subfigure}[b]{\textwidth}
    \centering
    \includegraphics[width=\textwidth]{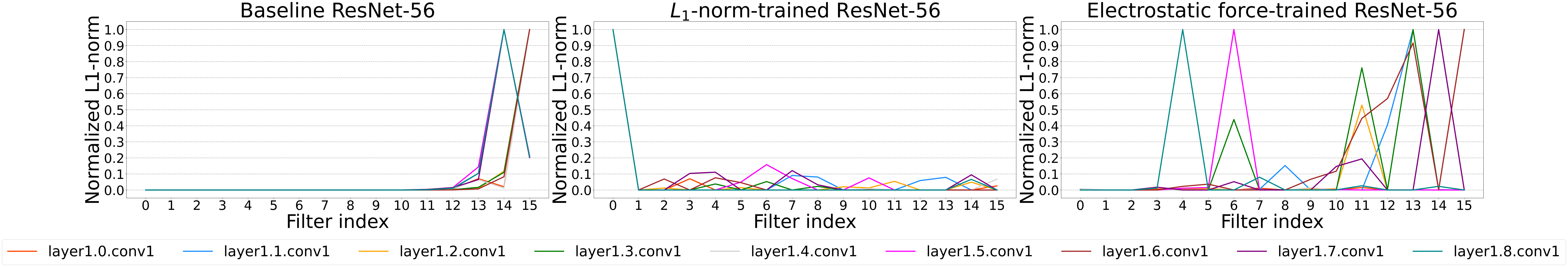}
  \end{subfigure} \\
  \begin{subfigure}[b]{\textwidth}
    \centering
    \includegraphics[width=\textwidth]{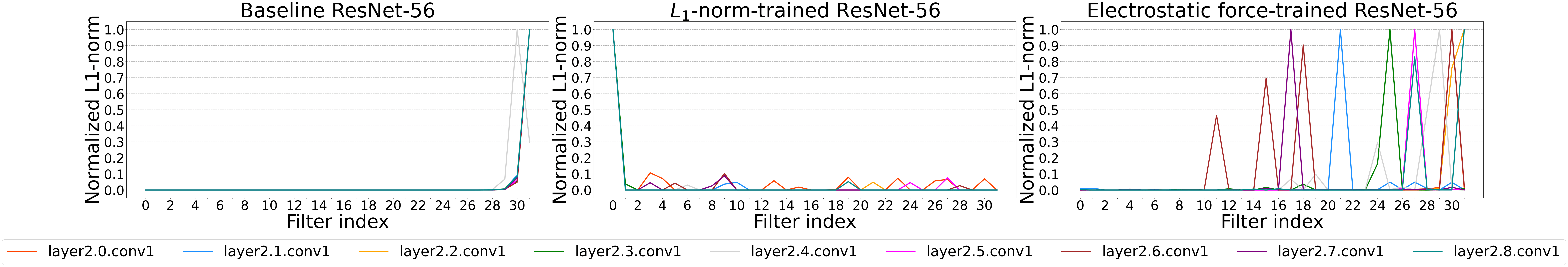}
  \end{subfigure}
  \caption{Normalized filter L$_1$-norm of layers 1 and 2 for the ResNet-56/CIFAR-10 model trained with and without electrostatic force, and with L$_1$-norm.}
  \label{figure3}
\end{figure*}

\subsection{Theoretical Analysis}
To theoretically explain the results obtained, we refer to Eq. \ref{eq11}. The penalty term added to the gradient term is due to the electrostatic force. When this force is attractive, the weights maintain non-zero values, and move toward zero when the force is repulsive. By pruning the filters with zero weights from the model, we accelerate the model while preserving information from attractive filters. Figure \ref{figure3} illustrates the effect of L$_1$-norm and the electrostatic force on the weights of filters (we only plotted convolutional layers to prune) in ResNet-56/CIFAR-10. In the baseline model, we can see that only some filters have their normalized L$_1$-norms different from zero. However, when training the model with L$_1$-norm, the number of filters with normalized L$_1$-norms different from zero are increased. Furthermore, when training with the electrostatic force, the number of filters with normalized L$_1$-norms different from zero remains significant. Nevertheless, these norms exhibit greater values. Consequently, when we prune filters with insignificant (or zero) normalized L$_1$-norms from the electrostatic force-trained model, we preserve the information from filters with significant (or non-zero) normalized L$_1$-norms. This analysis leads to the conclusion that our electrostatic force-based training method can be used to optimally configure both modern deep networks with residual connections and single-branch architectures for the pruning stage with minimal loss in accuracy.

\subsection{Comparison With State-of-the-arts}
The subsequent analysis will focus on the model accuracy after pruning and the acceleration achieved, which is quantified by the reduction in FLOPs. This is consistent with our objective of proposing a SP method that can be used to create an accelerated model with high accuracy. 

\subsubsection{CIFAR}
In general, our electrostatic force (p), despite exhibiting larger speedup rates, yields superior top-1 accuracy after pruning compared to all existing methods. For example, (1) with ResNet-56/CIFAR-10, our electrostatic force outperforms Torque by 0.12\% accuracy at the 2.17$\times$ speedup, while at 2.73$\times$, electrostatic force is better by 0.31\%. (2) With VGG-19/CIFAR-100, our electrostatic force outperforms Kron-OBD/OBS by 8.3/8.34\% accuracy at the 6.85$\times$ speedup. At the 8.89$\times$ speedup, our electrostatic force is better by 1.66\% accuracy compared to Torque. GReg-1/2 exhibits a slight superiority of 0.02/0.22\% accuracy over our electrostatic force. However, it is also important to note that the computational cost of GReg-1/2 is significantly higher than that of electrostatic force. Moreover, any modifications to the pruning rate in GReg-1/2 necessitate a complete retraining, a requirement that is not applicable in electrostatic force. This renders the electrostatic force more flexible and cost-effective.

\subsubsection{ImageNet}
As with the CIFAR results (Tables \ref{table4} and \ref{table5}), our method generally provides a good balance between maintaining high accuracy after pruning and achieving a high speedup (i.e., accuracy-FLOPs trade-off). In terms of pruned top-1 accuracy, our method, even with a larger speedup, achieves better or comparable performance compared to SOTA methods. For example, our method outperforms Taylor-FO by 0.89\% accuracy and GReg-1/2 by 0.18/0.11\% accuracy.

As observed from Tables \ref{table4}, \ref{table5}, and \ref{table6}, an electrostatic force-trained model using pretrained weights consistently achieved higher accuracy than a model trained with randomly initialized weights. This can be attributed to the fact that pretrained weights provide an optimal starting point (good minima), leading to better convergence during electrostatic force training.

\subsection{Ablation Study}
Our method relies heavily on the electrostatic force rate, which is denoted as $\alpha_e$ and serves as a crucial hyperparameter. This hyperparameter controls the intensity of the electrostatic force exerted on the filters in the convolutional layers of the model (as per Eq. \ref{eq7}). A substantial penalty is added to the gradient when $\alpha_e$ is large, resulting in significant variation in the updated weights. Conversely, selecting smaller values of $\alpha_e$ results in a minimal penalty to the gradient, which corresponds to a normal variation in the updated weights. 

By choosing an appropriate value of $\alpha_e$, a significant number of filters can have zero, enabling us to achieve higher pruning ratios without compromising the model's accuracy. To determine the most suitable value of $\alpha_e$, we trained the models with three distinct values of $\alpha_e$: $10^{-11}$, $10^{-12}$, and $10^{-14}$. Figure \ref{figure2} shows that the appropriate values of $\alpha_e$ for the ResNet-56 and VGG-19 models are $10^{-11}$ and $10^{-12}$, respectively. Based on this observation, we conclude that larger FLOPs models require small electrostatic force than those with smaller FLOPs to achieve the desired pruning results. 

\section{Conclusion and Future Work}
We proposed a novel method that integrates the concept of electrostatic force from physics into the training stage of DCNNs through regularization. We applied the electrostatic force to the convolution filters, either attracting or repulsing their weights toward non-zero or zero values, respectively. This resulted in a sparse model where the sparser part was constituted by filters that experienced the repulsing force and the denser part was represented by those that experienced the attracting force. This weight distribution allows for the model to be pruned by eliminating the repulsive filters, which represent less important weights, while preserving the information from the attractive filters, which represent more important weights. Our method demonstrated promising pruning results, with performance comparable to SOTA SP methods. In the future, We plan to propose novel methods that utilize electrostatic force concept from physics to address challenges in model pruning.


\end{document}